\documentclass[10pt,twocolumn,letterpaper]{article}

\usepackage{cvpr}      

\usepackage{graphicx}
\usepackage{amsmath}
\usepackage{amssymb}
\usepackage{booktabs}

\usepackage{graphicx}
\usepackage{wrapfig}
\usepackage{subcaption}
\usepackage[ruled,vlined]{algorithm2e}

\usepackage{amsthm}
\usepackage{url}

\usepackage[pagebackref,breaklinks,colorlinks]{hyperref}

\usepackage[capitalize]{cleveref}
\crefname{section}{Sec.}{Secs.}
\Crefname{section}{Section}{Sections}
\Crefname{table}{Table}{Tables}
\crefname{table}{Tab.}{Tabs.}

\begin{document}
\title{Locally Asynchronous Stochastic Gradient Descent for Decentralised Deep Learning}
\author{Tomer Avidor \and Nadav Israel}

\author{Tomer Avidor\\
Edgify.ai\\
{\tt\small tomeravidor@yahoo.com}
\and
Nadav Israel\\
Edgify.ai\\
{\tt\small nadav.israel@edgify.ai}
}

\maketitle

\begin{abstract}
Distributed training algorithms of deep neural networks show impressive convergence speedup properties on very large problems. However, they inherently suffer from communication related slowdowns and communication topology becomes a crucial design choice. Common approaches supported by most machine learning frameworks are:
1) Synchronous decentralized algorithms relying on a peer-to-peer All Reduce topology that is sensitive to stragglers and communication delays.
2) Asynchronous centralised algorithms with a server based topology that is prone to communication bottleneck. Researchers also suggested asynchronous decentralized algorithms designed to avoid the bottleneck and speedup training, however, those commonly use inexact sparse averaging that may lead to a degradation in accuracy. 
In this paper, we propose Local Asynchronous SGD (LASGD), an asynchronous decentralized algorithm that relies on All Reduce for model synchronization.

We empirically validate LASGD's performance on image classification tasks on the ImageNet dataset. Our experiments demonstrate that LASGD accelerates training compared to SGD and state of the art  gossip based approaches.
\end{abstract}

\section{Introduction}

\begin{figure}[t]
    \centering
\begin{subfigure}{.5\columnwidth}
   \captionsetup{width=1\linewidth}
    \includegraphics[width=\columnwidth]{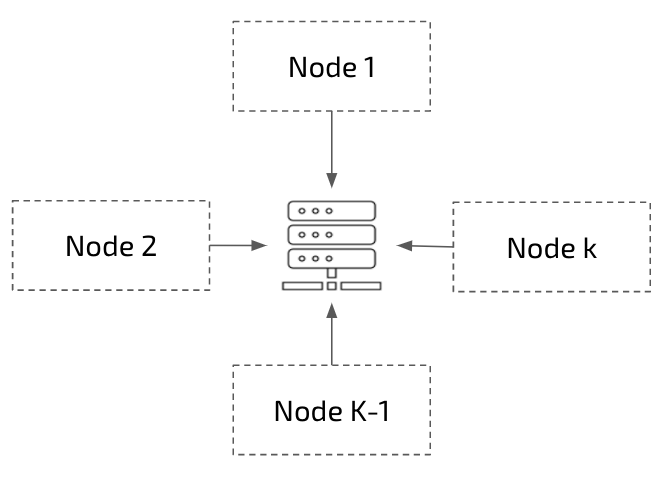} 
    \caption{Centralized.}
    \label{fig:centralized}
\end{subfigure}
\begin{subfigure}{.5\columnwidth}
    \captionsetup{width=1\linewidth}
    \includegraphics[width=\columnwidth]{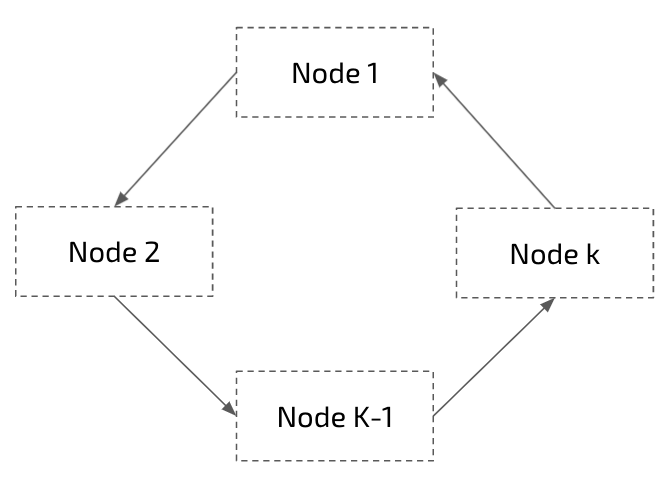}\caption{Decentralized.}
    \label{fig:decentralized}
\end{subfigure}
\caption{Common distributed topologies. (a) Centralized topology uses a server to synchronize the model updates from all nodes. (b) Decentralized topology uses peer to peer communication to synchronize model updates.}
\end{figure}

Deep learning achieves state of the art results in many machine learning tasks. Each year new techniques are presented superseding the current best algorithms. However, the time it takes to train a model to convergence is still far from optimal and many applications straggle to obtain state of the art results while reducing training duration. 

In recent years, real world applications are raising and motivating work on model architecture optimization producing models that train faster at the expanse of lower accuracy such as: MobileNetV2 ~\cite{MobileNetV2}, ShuffleNetV2 ~\cite{ShuffleNetV2}, and CondenseNet ~\cite{Condensenet}. However, when training on large data sets such as ImageNet ~\cite{imagenet} training time is still long leading to high costs. Distributed training is the common way to reduce training time, however, it often leads to higher costs as communication between training nodes slows down the training process.

\par
\begin{figure}
    \centering
    \includegraphics[width=0.3\textwidth]{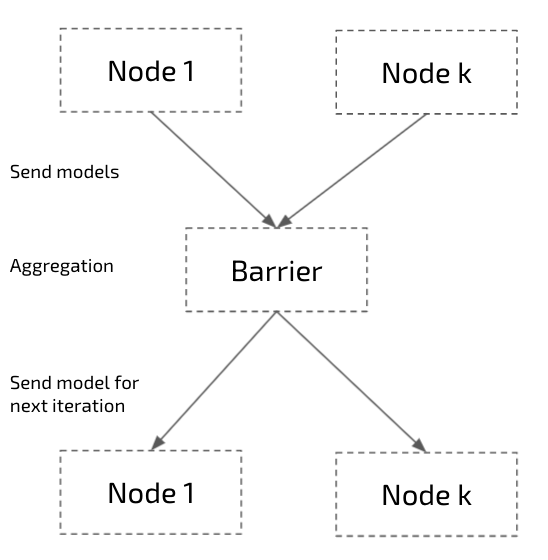}
\caption{Synchronization step in distributed synchronous training.}
        \label{fig:synchronous}
\end{figure}

 Large mini-batch distributed SGD is the current workhorse ~\cite{ImageNet15Minutes, Goyal2017AccurateLM} for training on large datasets. Modern deep-learning frameworks such as TensorFlow ~\cite{tensorflow2015-whitepaper} MXNet ~\cite{Chen2015MXNetAF} and PyTorch ~\cite{pytorch} support distributed training through both Decentralized and Centralized topologies depicted in ~\cref{fig:centralized} for a Centralized topology and in ~\cref{fig:decentralized} for a Decentralized topology. In these methods, as shown in ~\cref{fig:synchronous}, nodes compute local mini-batch gradients and then aggregate all local gradients to an average global gradient at the Barrier/Synchronization stage. The aggregation is typically done with an All-Reduce primitive ~\cite{10.1177/1094342005051521, 10.1007/978-3-540-24685-5_1}. However, the synchronous nature of All-Reduce hampers scaling and is not as robust to node heterogeneity as asynchronous methods. On the other hand, asynchronous algorithms typically use a central server ~\cite{10.5555/2999134.2999271, 10.5555/2968826.2968829, 10.5555/2685048.2685095} for aggregation which may become a bottleneck in communication and slow down convergence. On top of that, typical modern data centers such as AWS and Azure offer symmetric resources in the sense that all computational nodes have the same bandwidth and so communication is optimally utilized by All-Reduce algorithms such as Ring-All-Reduce ~\cite{10.1177/1094342005051521}. 

\par This motivates the study of algorithms for asynchronous training in a decentralized (All-Reduce) fashion. The prevalence of distributed training resulted in many efficient implementations of All-Reduce algorithms such as NCCL ~\cite{nccl}, MPI ~\cite{MPI} and Horovod ~\cite{Horovod}. There is a large body of work on asynchronous decentralized algorithms ~\cite{MALT, SGP, ADPSGD} but those focus on sparse graph communication and do not take advantage of the efficiency of All Reduce implementations. 

In this work we present Local Asynchronous SGD (LASGD), an algorithm breaching the gap between efficient decentralized All-Reduce and asynchronous training. In LASGD nodes do not wait for other nodes and communicate asynchronously. By overlapping the communication with the computation, we enable nodes to communicate through a synchronous All-Reduce while computing the next mini-batch gradient asynchronously allowing both high accuracy and faster training. We adopt Elastic Averaging ~\cite{EASGD} to allow nodes to run several local mini-batch iterations while the All-Reduce computation runs in parallel. LASGD achieves linear speedups with respect to the number of nodes while maintaining accuracy on par with SGD. We conduct experiments on image classification on the ImageNet dataset on AWS and show competitive results with up to 32 nodes each with 4 Tesla V100 GPUs.    

\section{Related Work}
Many approaches have been studied to accelerate distributed training. In this section we review those with respect to their communication topology.

\paragraph{Decentralized Synchronous}
All-Reduce implementations of SGD have been used extensively and are common place. In these methods, hundreds of nodes train collectively. At each iteration each node fetches the model, computes a local mini-batch gradient, aggregates all the local gradients using All-Reduce and updates the model.
In Goyal \etal ~\cite{Goyal2017AccurateLM} the authors trained a model on the ImageNet data set using hundreds of GPUs. All-Reduce was done using a recursive halving doubling algorithm ~\cite{10.1007/978-3-540-24685-5_1, 10.1177/1094342005051521}. However, for the training to run in the given time of one hour the authors used a 50GB network card, unfortunately common cloud provider often do not offer such network bandwidth (E.g. AWS p3.8xl only comes with a 10GB card). 

Other papers ~\cite{DBLP:journals/corr/LanLZ17, DistributedSubgradientProjection, D2} suggested reducing the number of inter node communications by connecting nodes sparsely via a connected graph $G$. Lian \etal ~\cite{10.5555/3295222.3295285} provide a theoretical analysis indicating that decentralized algorithms can outperform centralized algorithms. In this algorithm each node sends $O(deg(G))$ times the model size for each communication as apposed to $O(1)$ for All-Reduce. The accuracy of the algorithm  suffers when $deg(G)$ is reduced.

\paragraph{Asynchronous Centralized}
Parameter sever methods ~\cite{10.5555/2999134.2999271, 10.5555/2968826.2968829, 10.5555/2685048.2685095} employ a centralized topology where nodes asynchronously pull a model from the server, compute a local gradient on a mini-batch and push an update to the server. The synchronization step constitutes an idle time for the nodes and a server communication bottleneck might slow down convergence in wall clock time. 

In Zhang \etal ~\cite{EASGD} the authors present \textit{Elastic Averaging SGD (EASGD)} where each node trains a local model that is constrained to the global model by a quadratic term added to the loss. This method is closest to our method in that LASGD also relies on the quadratic (elastic averaging) term for the global update only LASGD is decentralized where EASGD uses a centralized server topology. In EASGD nodes run a constant number of mini-batch gradient computations between consecutive global updates and the updates are done in a round robin fashion similar to ~\cite{10.5555/2984093.2984354}. The difference from LASGD is that in LASGD nodes synchronize dynamically as every node keeps computing local gradients until the All-Reduce ends leading to less idle time and better utilization of the data-center infrastructure. We found that although cloud providers offer symmetric nodes (i.e. all nodes have the same computation and communication properties) gradient computation time is quite heterogeneous with some nodes taking twice as much time to compute a mini batch gradient. Allowing each node to compute a different number of local mini batch gradients, LASGD is able to adapt to to these variations.

\paragraph{Asynchronous Decentralized}
Li \etal ~\cite{10.5555/3327757.3327900}  present a pipe-line implementation of ~\cite{10.5555/2984093.2984354} that reduces node idle time by computing a new mini-batch gradient during (in parallel) model synchronization. The effectiveness of the method depends on the ratio between computation to synchronization time and is optimal when the ratio is 1.

Another popular approach presented in Lian \etal ~\cite{ ADPSGD} and Zhanhong \etal ~\cite{10.5555/3295222.3295340} relies on gossip-based communication to propagate the model between nodes. Nodes communicate on a time varying sparse graph. Similar to the synchronous case, on each synchronization round each node sends $O(deg(G))$ the model size. These methods  require  symmetric communication (that is if node $i$ sends to node $j$ it must also receive from node $j$ before proceeding) and so need some deadlock avoidance mechanism making them slower. In Assran ~\etal ~\cite{SGP} the authors utilize the PushSum ~\cite{1238221, 7405263} algorithm to avoid symmetric communication and speedup training. Similar to our method, they also compute local gradients in parallel to communication, however, their communication module is not collaborative and requires a higher bandwidth of $O(deg(G))$ times the model size for each node when Ring-All-Reduce uses $O(1/P)$ times the model size where $P$ is the number of nodes.

\section{Problem setting}
\newcommand\suml{\sum\limits}
\newcommand\barx{\bar{x}}
\newcommand\sumg{\frac{1}{P}\sum\limits_{i=1}^P}
\newcommand\sumlg{\suml_{t=0}^{\tau_1}g_t^i}
\newcommand\sumlgg{\suml_{t=\tau_1}^{\tau_2}g_t^i}
\newcommand\sumlgj{\suml_{t=\tau_{j-1}}^{\tau_j}g_t^i}
\newcommand\numeq[1]%
  {\stackrel{\scriptscriptstyle(\mkern-1.5mu#1\mkern-1.5mu)}{=}}
\newcommand\simeql[1]%
  {\stackrel{\scriptscriptstyle(\mkern-1.5mu#1\mkern-1.5mu)}{\simeq}}
\newcommand\eob{\frac{\eta}{\beta}}
\newcommand\sumln{\suml_{i=1}^P}
\newcommand\sump{\sum\limits_{i=1}^P}

Consider the problem of minimizing a function $F(x)$ in a distributed computation environment ~\cite{10.5555/59912} with $P$ computational nodes. We look at the following optimization problem: \[ \min_x F(x) = \min_x \mathbb{E}_{\zeta \backsim \mathcal{D}}(f(x,\zeta)) \]
 Where $\mathcal{D}$ is a distribution over the data and $\zeta$ is sampled from $D$. The objective can be split by dividing the data into $P$ parts $D_i$ : 
 \[\min_x \sumln F_i(x) =  \min_x \sumln \mathbb{E}_{\zeta_i \backsim \mathcal{D}_i}(f(x,\zeta_i)) \] 
 An equivalent formalization for the distributed setting is the \emph{global consensus problem} ~\cite{ADMM, OptimizationTheory}:
\begin{equation}
\begin{aligned}
    \min \sumln F_i(x^i) \\
    \text{subject to} \quad x^i=z
\end{aligned}
\end{equation}
This optimization problem can be reformulated as: 
\begin{equation} \label{eq:obj}
    \begin{aligned}
        \min \sumln F_i(x^i) + \frac{1}{2}\rho  \Vert x^i -z \Vert^2        
    \end{aligned}
\end{equation}
Following ~\cite{EASGD} using gradient descent on ~\Cref{eq:obj}:

\begin{subequations}
    \begin{align}
        x_{t+1}^i=x_t^i - \eta_t (g_t^i + \rho(x^i_t-z_t)) \\
        z_{t+1}=z_t+\gamma_t \sump(x^i_t-z_t) \label{eq:z1}
    \end{align}
\end{subequations}
By denoting $/betta_t=P /gamma_t$  \Cref{eq:z1} becomes:

\begin{equation}
z_{t+1}=z_t+\beta_t \sumg (x_t^i-z_t)
\end{equation}
 or equivalently:
\begin{equation}
z_{t+1}=(1-\beta_t)z_t+\gamma_t \sumg x_i^t \label{eq:synch_z}
\end{equation}
Similarly by taking $\alpha=\eta_t\rho$:
\begin{equation}
x_{t+1}^i=\alpha z_t + (1-\alpha) x_t^i - \eta_t g_t^i \label{eq:synch_x}
\end{equation}
The convergence properties for this algorithm were discussed in Zhang \etal ~\cite{EASGD}. We look at a special case of this algorithm that for the asynchronous case described in  \cref{sec:LocallyAsynchronousSGD} gave better overall results (on the experiments described in \cref{sec:experemints}).

For the case where $\beta=1$ we get that $z_{t+1} = \sumg x_t^i$ the average of the local models. 
When $\alpha=\beta=1$ we get the simple update:
\begin{align}
x_{t+1}^i= z_t - \eta_t g_t^i \label{eq:local_update} \\
z_t=\sumg x_t^i 
\end{align}

\begin{figure}[t]
    \centering
    \includegraphics[width=7cm]{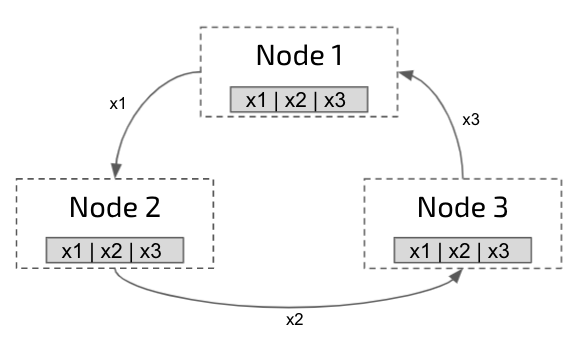}
    \caption{Ring all reduce with 3 nodes. At every iteration, each nodes sends a third of the data, making the peak bandwidth the size of the data.}
    \label{fig:RAR}
\end{figure}

\section{Locally Asynchronous SGD}
\label{sec:LocallyAsynchronousSGD}
In the previous section we presented the optimization updates for the synchronous case. In this section we will focus on the asynchronous algorithm. 

\textbf{Communication Topology:} To avoid using an additional server and the communication bottleneck of centralized distribution, LASGD is decentralized and uses Ring All Reduce ~\cite{10.1177/1094342005051521} for optimal bandwidth usage. Ring All Reduce is an iterative algorithm where each of the $P$ nodes 
sends $|x| / P$ bits every iteration with $|x|$ being the model size. At every iteration each node sends one chunk to its neighbor in the ring. ~\Cref{fig:RAR} depicts the first iteration in ring reduce with 3 nodes. This ensures a constant low bandwidth consumption the size of $|x|/P$ per node with the trade-off being a higher latency. Thus, LASGD ensures low bandwidth consumption. In contrast, SGP ~\cite{SGP} synchronizes through a broadcast algorithm where the bandwidth is $|x|$ for each node.

 \textbf{Asynchronous Algorithm:} The algorithm's pseudo code is shown in ~\cref{alg:LASGD}. The key point is that the nodes compute a synchronous All Reduce while overlapping it with mini-batch gradient computations. Every node keeps a local clock $t^i$ which is incremented every gradient update and a global clock $t$ that is updated every global update.  
 All nodes start training with the same random model. At the first iteration, each node computes a mini batch gradient $g^i_0$ and then performs All-Reduce on $g^i_0$. Without waiting for the All-Reduce to end, the nodes compute mini-batch gradients and update their local model until the All-Reduce ends or a threshold on the number of local model updates is met. The algorithm thus dynamically adapts the global synchronization rate (that is, the number of local-gradient update steps between consecutive All-Reduce steps) leading to optimizing GPU utility. Once All-Reduce is done, each local node (asynchronously from each other) updates its local model as in \cref{alg:global_update} essentially adding the sum of local gradients to the average model. The nodes then send the updated local model to the global All-Reduce as in \cref{alg:global_ar} and continue to compute the next mini-batch gradient.

\begin{algorithm}[h]
\SetAlgoLined
\SetKwInOut{Input}{input}\SetKwInOut{Output}{output}

\Input{Initial model weights $x^i_0$, learning rate $\eta_t^i$ and  $\tau_{max}$}
\SetAlgoLined
{
\begin{enumerate}
    \item $x^i \leftarrow x^i_0$
    \item Compute mini-batch gradient: $g^i_0$
    \item $x^i \leftarrow x^i - \eta_0 g^i_0$
    \item Send $x^i$ to All Reduce.
    \item $\tau_i=0$
\end{enumerate}
}
\For{$t^i \in {1 \dots T}$ }
{
\begin{enumerate}
\setcounter{enumi}{5}
\item Compute mini-batch gradient $g^i(x^i)$
\item $x^i \leftarrow x^i-\eta_{t^i} g^i(x^i)$
\item $\tau_i = \tau_i +1$
\item If All Reduce done or $\tau_i == \tau_{max}$: 
\begin{enumerate}
    \item $x^i_{t+1}=z_t+(x^i-x^i_{t})$ \label{alg:global_update}
    \item $x^i \leftarrow x^i_{t+1}$
    \item Send $x^i_{t+1}$ to All-Reduce. \label{alg:global_ar}
    \item $t = t+1$
    \item $\tau_i = 0$
\end{enumerate}

\end{enumerate}
}
\caption{LASGD}
\label{alg:LASGD}
\end{algorithm}

\section{Empirical Results}
\label{sec:experemints}
We experimentally compare LASGD with All-Reduce SGD and with SGP ~\cite{SGP}. For SGP we used the code provided by the authors in ~\cite{SGP_code}. SGP has a number of formulations that give different accuracy to run-time trade-offs, in this paper when we say SGP we refer to overlap SGP with 1-Peer topology (1-OSGP).

\subsection{Experiment Methodology}

\subsubsection{Datasets and Models}
We run experiments on the ImageNet-1k ~\cite{imagenet} data set and use PyTorch ~\cite{pytorch} for our deep learning framework. We use NVIDIA NCCL All Reduce collectives ~\cite{nccl} for the communication scheme. We trained Resnet50 ~\cite{ResNet}, which has a model size of 100MB, on ImageNet.

\subsubsection{Hardware}
All experiments were run on AWS instances. We evaluated LASGD on p3.8xlarge nodes. Each node has 4 NVIDIA Tesla V100 GPUs, 32  Intel Xeon E5-2686 v4 vCPUS and 10Gbit/s Ethernet. We use up-to 32 nodes (128 GPUs) for an experiment.

\begin{table*}[t]
\begin{center}
    \begin{tabular}{c c c c}
    \hline
        Number of Nodes & 8 (32 GPUs) & 16 (64 GPUs) & 32 (128 GPUs)\\
    \hline
        SGD-AR & 76.2\% \hspace*{0.1cm} (5.75 hr.) & 76.1\% \hspace*{0.1cm} (3.25 hr.) & 76\% \hspace*{0.1cm} (2.02 hr.) \\
        SGP    & 75.8\% \hspace*{0.1cm} (5.45 hr.) & 75.5\% \hspace*{0.1cm} (3.15 hr.) & 74.4\% \hspace*{0.1cm} (2.1 hr.) \\
    \hline
        WLASGD  & 75.8\% \hspace*{0.1cm} (4.37 hr.) &  75.1\% \hspace*{0.1cm} (2.2 hr.) & 72.8\% \hspace*{0.1cm} (1.17 hr.) \\
        \textbf{LASGD} & 75.9\% \hspace*{0.1cm} (4.77 hr.) &	75.8\% \hspace*{0.1cm} (2.55 hr.)& 75\% \hspace*{0.1cm}(1.3 hr.) \\

    \hline
    \end{tabular}
    \caption{Top 1 validation accuracy (\%) and run time (in hours) for the ImageNet dataset for SGD All-Reduce, SGP and LASGD. In WLASGD every GPU is a node, in LASGD and SGP every VM (4GPUs) constitutes a single node.}
    \label{table:1}
\end{center}
\end{table*}

\subsection{Imagenet}
\label{sec:imagenet}
We trained a Resnet50 on the ImageNet-1k dataset. We examine scaling by running on 8, 16 and 32 nodes (having 32, 64 and 128 GPUs respectively). We follow the training procedure in Goyal \etal ~\cite{Goyal2017AccurateLM} only we use a batch size of 64 images per GPU. Every node has a total batch size of 256. Following Goyal \etal ~\cite{Goyal2017AccurateLM} we warmup the learning rate for the first 5 epochs to $0.1*P$ where $P$ is the number of nodes, we train for 90 epochs and the learning rate is decayed by a factor of 10 at epochs 30, 60 and 80. We use Nestrov momentum and weight decay. 

We examine 2 setups: 
\textbf{A node setup,} used in both SGP and LASGD. Where (using NVIDIA NCCL All-Reduce) each node is a single worker, effectively increasing each worker's batch size to 256 thus providing better stability and improved convergence properties.
\textbf{A worker setup,} implemented solely for LASGD. Where as each GPU is assigned a single worker leading to reduced communication demands and faster run time, we call this setup WLASGD.

In both setups for LASGD, $\tau_{max}$ was set to 1 to maximize accuracy.

~\Cref{table:1} shows the total training time and the validation accuracy for the various setups. For any number of computational nodes, both WLASGD and LASGD out perform SGD and SGP in terms of run time. We observe that SGP's run time is similar to SGD's and deteriorates as the number of node increases. We hypothesize that this slow down is because SGP communicates model weights during the Push phase using broadcast directive which uses higher bandwidth. Further, we observe that using exact All-Reduce in LASGD results in better accuracy compared to SGP and a milder degradation compared to SGD as the number of nodes increases.

\begin{figure*}%
    \centering
    \subfloat[\centering 8 Nodes (32 GPUs)]{{\includegraphics[width=8cm]{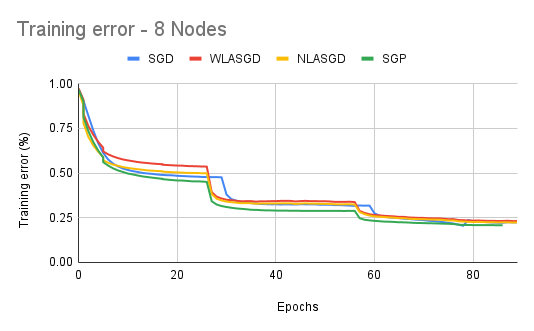} }}%
    \qquad
    \subfloat[\centering 16 Nodes (64 GPUs)]{{\includegraphics[width=8cm]{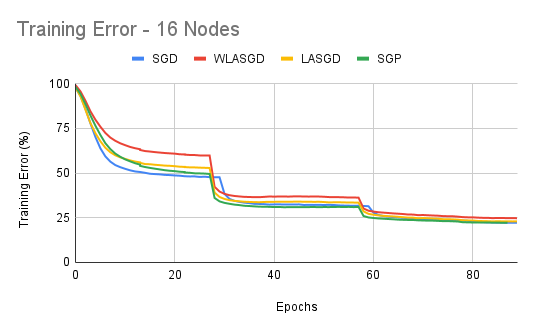} }}%
    \subfloat[\centering 32 Nodes (128 GPUs)]{{\includegraphics[width=8cm]{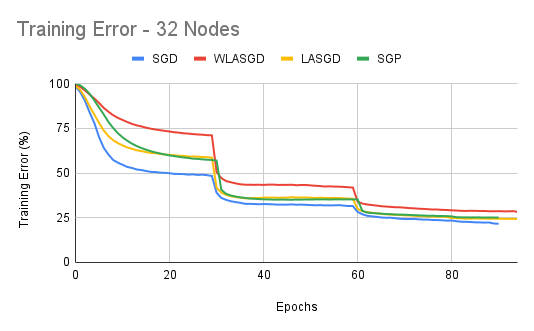} }}%
    \caption{Training error by epoch on ImageNet of SGD, LASGD, WLASGD (LASGD where every GPU is a node) and SGP. Results are presented by the number of nodes. Models were trained for 90 epochs.}%
    \label{fig:imgnet_epoch}%
\end{figure*}

\begin{figure*}%
    \centering
    \subfloat[\centering 8 Nodes (32 GPUs)]{{\includegraphics[width=8cm]{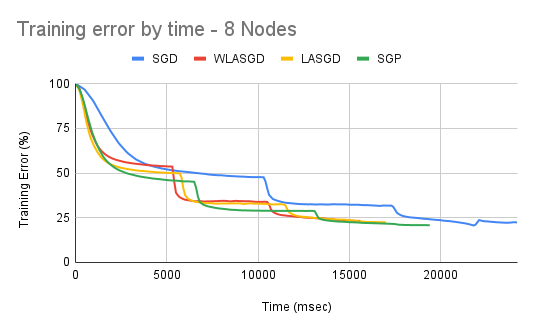} }}%
    \qquad
    \subfloat[\centering 16 Nodes (64 GPUs)]{{\includegraphics[width=8cm]{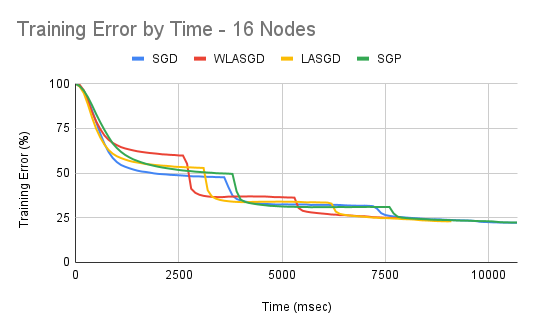} }}%
    \subfloat[\centering 32 Nodes (128 GPUs)]{{\includegraphics[width=8cm]{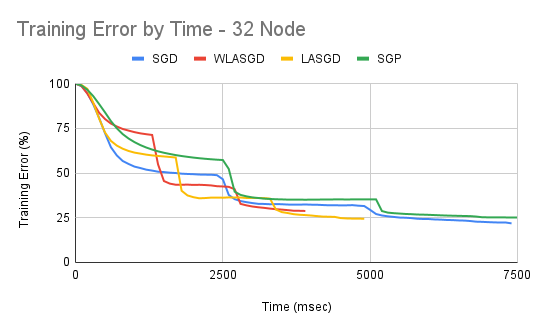} }}%
    \caption{Training error by wall clock time on ImageNet of SGD, LASGD, WLASGD (LASGD where every GPU is a node) and SGP. Results are presented by the number of nodes. Models were trained for 90 epochs. The results show that the benefit from LASGD and WLASGD increases with the number of nodes.}%
    \label{fig:imgnet_time}%
\end{figure*}

~\Cref{fig:imgnet_epoch} shows the training error of SGD, SGP, LASGD and WLASGD by epoch on 8, 16 and 32 nodes. ~\Cref{fig:imgnet_time} shows training error by time, in all cases LASGD computes 90 epochs in less time than SGP and SGD.

\begin{figure}
    \centering
    \includegraphics[width=8cm]{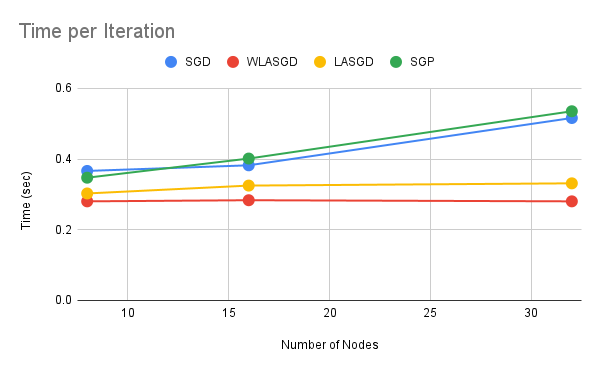}
    \caption{Scaling properties for SGD, SGP, LASGD and WLASGD on ImageNet when the number of nodes is increased from 8 to 16 and 32. LASGD and WLASGD show good scaling properties at this rage while SGP is on par with SGD and does not scale well.}
    \label{fig:time_per_iteration}
\end{figure}

~\Cref{fig:time_per_iteration} examines the scaling properties of the methods. The figure shows the average iteration time (computation and communication) for SGD, SGP, LASGD and WLASGD for 8, 16 and 32 nodes. For both LASGD and WLASGD the iteration time remains constant meaning they scale linearly with the number of nodes while SGP decays similarly to SGD.

\subsection{Communication Intensive}
In a communication intensive setting we expect that LASGD would improve run time further and that increasing $\tau_{max}$ would bring an additional reduction in run time. Our experiments showed that reducing each GPU's batch size to 32 increases the communication intensity (i.e. the number of communication steps per second) by $20\%$, it further releases some extra computational resources to compute the (overlapping) communication. We evaluated LASGD on the ImageNet data set as in \cref{sec:imagenet} on 16 nodes (64 GPUs) with batch size of 32 per node, the learning rate was changed to match ~\cite{Goyal2017AccurateLM}. In order to test the usefulness of increasing $\tau_{max}$ we set $\tau_{max}=5$ for each iteration and examined the number of mini batch gradient computations performed during a single All-Reduce. \Cref{table:tau5} shows that over $80\%$ of All-Reduce instances needed more than one gradient computation time to complete the All-Reduce and over $70 \%$ of the instances took between 1 and 2 gradient computations. Note that about $4\%$ of the instances did not end even after 5 gradient computations, we theorize that this is due to either communication error or process synchronization error, this can be handled by relaxing the condition on the number of nodes needed to start All-Reduce as described in \cref{sec:limitation}. The results presented in \Cref{table:comp_32_64} show that both SGD and LASGD performed better in terms of run time for the mini-batch 32 scenario than for the mini-batch size 64 scenario, as expected. LASGD was able to capitalize on setting $\tau_{max}=5$, having the fastest run time at the cost of reduced accuracy. The speedups are summarized in \Cref{table:speedups}.

\begin{table*}[t]
\begin{center}
    \begin{tabular}{c c c c c c c c}
    \hline
        Number of computed mini-batch gradients & 1 & 2 & 3 & 4 & 5 & 6+ \\
    \hline
       Number of instance in percentage  & 11.2 \% & 72.8\% & 11.1\% & 0.2\% & 0.6\%& 3.9\% \\
    \hline
    \end{tabular}
    \caption{The number of min-batch computations required to compute an All-Reduce in LASGD. The figures shown are the number (in \%) of iterations for which it took from 1 up to 6 mini-batch computations for the All-Reduce computation to end.}
    \label{table:tau5}
\end{center}
\end{table*}

\begin{table}[t]
\begin{center}
    \begin{tabular}{c c c c}
    \hline
        Mini-Batch Size & 32 & 64 \\
    \hline
        SGD-AR & 76.1\% \hspace*{0.1cm} (5.42 hr.) & 76.1\% \hspace*{0.1cm} (3.25 hr.) \\
        SGP    & 75.6\% \hspace*{0.1cm} (5.47 hr.) & 75.5\% \hspace*{0.1cm} (3.15 hr.) \\
    \hline
        LASGD ($\tau_{max}=1$) & 75.9\% \hspace*{0.1cm} (3.97 hr.) &	75.8\% \hspace*{0.1cm} (2.55 hr.) \\
        LASGD ($\tau_{max}=5$)  & 74.8\% \hspace*{0.1cm} (3.67 hr.) &  \\
    \hline
    \end{tabular}
    \caption{Comparison of top 1 validation accuracy (\%) and run time (in hours) for communication intensive setup (mini batch size 32) and computation intensive (batch size 64). ImageNet data set using 16 nodes (64 GPUs).}
    \label{table:comp_32_64}
\end{center}
\end{table}

\begin{table}[t]
\begin{center}
    \begin{tabular}{c c c c}
    \hline
        Mini-Batch Size & 32 & 64 \\
    \hline
        SGD-AR & 1 & 1 \\
        SGP    & 1.21 & 1.03 \\
    \hline
        LASGD ($\tau_{max}=1$) & 1.36 &	1.27 \\
        LASGD ($\tau_{max}=5$)  & 1.47 &  \\
    \hline
    \end{tabular}
    \caption{Speedup in time compared to SGD for the 32 and 64 batch size instances. Results show that in the communication intensive setup (batch size 32) SGP and LASGD had better speedups compared to the computation intensive setup.}
    \label{table:speedups}
\end{center}
\end{table}

\section{Limitations}
\label{sec:limitation}
LASGD relies heavily on the ability of deep learning frameworks and hardware to compute gradients while synchronizing model weights. We found support for this property in PyTorch ~\cite{pytorch} through the use of CUDA Streams. 

We note that in a heterogeneous system with stragglers (i.e. slower nodes) LASGD's synchronization rate is dictated by the straggler. By adapting the value of $\tau_{max}$ LASGD offers dynamic synchronization rates and avoids idle time by allowing faster node to compute more local gradients. A subject for future work is to allow partial All-Reduce. That is, start a new All-Reduce with only a fraction (say $80\%$) of the nodes. This way, the faster nodes will run more updates and better utilize their resources while slower nodes will synchronize at a slower rate.

The performance of LASGD (and indeed all distributed algorithms) depends greatly on the communication to computation ratio, that is the ratio between the time it takes to run All Reduce to the time it takes compute mini batch gradient. This ratio, in turn, is dependant on various factors such as: band width, latency,  network topology, computation power, model size, model complexity and so on. As a result, the benefit from using LASGD is difficult to predict and has to be empirically calculated on a case by case basis. 

We note that for the synchronous case of \cref{eq:synch_z} and \cref{eq:synch_x} Zhang \etal ~\cite{EASGD} prove convergence for a strongly convex objective for $0 \leq \gamma \leq 1$ and $0 \leq \alpha < 1$. It has been our experience that for the asynchronous case of \cref{sec:LocallyAsynchronousSGD} the update rule presented in \cref{eq:local_update} where we use $gamma=alpha=0$ proved superior in term of accuracy.
 
 \section{Conclusion}
This paper presents LASGD, an asynchronous decentralized algorithm based on Elastic Averaging and Ring All Reduce. We empirically study the usefulness of the method by training on ImageNet using several computational setups. We show that LASGD significantly reduces computation time compared to SGD.  

\clearpage
{\small
\bibliographystyle{ieee_fullname}
\bibliography{references}
}
\end{document}